\pdfoutput=1

\documentclass[11pt]{article}

\usepackage[final]{acl}

\usepackage{times}
\usepackage{latexsym}
\usepackage{amssymb}
\usepackage{amsmath}
\usepackage{algorithm}
\usepackage{algorithmicx}
\usepackage{algpseudocode}
\usepackage{tabularx}
\usepackage{booktabs}
\usepackage{multicol}
\usepackage{multirow}
\usepackage{xcolor}

\usepackage[T1]{fontenc}

\usepackage[utf8]{inputenc}

\usepackage{microtype}

\usepackage{graphicx}

\title{Semi-Supervised Formality Style Transfer with Consistency Training}

\author{Ao Liu,  An Wang, Naoaki Okazaki \\
      Tokyo Institute of Technology \\ 
      \texttt{liu.ao@nlp.c.titech.ac.jp} \\
      \texttt{wang@de.cs.titech.ac.jp} \\
      \texttt{okazaki@c.titech.ac.jp}
      }

\begin{document}
\maketitle

\begin{abstract}

Formality style transfer (FST) is a task that involves paraphrasing an informal sentence into a formal one without altering its meaning. To address the data-scarcity problem of existing parallel datasets, previous studies tend to adopt a cycle-reconstruction scheme to utilize additional unlabeled data, where the FST model mainly benefits from target-side unlabeled sentences. In this work, we propose a simple yet effective semi-supervised framework to better utilize source-side unlabeled sentences based on consistency training. Specifically, our approach augments pseudo-parallel data obtained from a source-side informal sentence by enforcing the model to generate similar outputs for its perturbed version. Moreover, we empirically examined the effects of various data perturbation methods and propose effective data filtering strategies to improve our framework. Experimental results on the GYAFC benchmark demonstrate that our approach can achieve state-of-the-art results, even with less than 40\% of the parallel data\footnote{Code available at \url{https://github.com/Aolius/semi-fst}.}.

\end{abstract}

\section{Introduction}
Formality style transfer (FST)~\cite{rao2018dear} has garnered growing attention in the text style transfer community, which aims to transform an \textit{informal}-style sentence into a \textit{formal} one while preserving its meaning. The large amount of user-generated data from online resources like tweets often contain informal expressions such as slang words (e.g., \textit{gonna}), wrong capitalization or punctuations, and grammatical or spelling errors. FST can clean and formalize such noisy data, to benefit downstream NLP applications such as sentiment classification~\cite{cari}. Some examples of FST data are presented in Table \ref{tab:example}.

\begin{table}[t]
    \centering
    \resizebox{\columnwidth}{!}{
    \begin{tabular}{|c|l|}
        \hline
       Informal  & \textit{TITANIC I THINK IT COST ABOUT 300 MILLION }\\ 
       Formal  &  \textit{I think that Titanic cost around 300 million dollars. }\\ \hline
       Informal &\textit{ being condiderate of her feelings and needs} \\ 
       Formal &\textit{ I am being considerate of her personal needs and feelings.} \\
       \hline
    \end{tabular}}
    \caption{Examples of informal-formal sentence pairs.}
    \label{tab:example}
\end{table}

With the release of the FST benchmark Grammarly Yahoo Answers Corpus (GYAFC) \citep{rao2018dear}, previous studies on FST tend to employ neural networks such as sequence-to-sequence (seq2seq) models to utilize parallel (informal and formal) sentence pairs. However, GYAFC only contains 100k parallel examples, which limits the performance of neural network models. Several approaches have been developed to address the data-scarcity problem by utilizing unlabeled sentences. In a previous study, \citet{zhang2020parallel} proposed several effective data augmentations methods, such as back-translation, to augment parallel data. Another line of research \citep{shang2019semi, xu2019formality, chawla-yang-2020-semi} conducted semi-supervised learning (SSL) in a cycle-reconstruction manner, where both forward and backward transfer models were jointly trained while benefiting each other by generating pseudo-parallel data from unlabeled sentences. Under this setting, both additional informal and formal sentences are utilized; however, the forward informal$\to$formal model mostly benefits from the \textit{target}-side (formal) sentences, which are back-translated by the formal$\to$informal model to construct pseudo training pairs. Conversely, the formal$\to$informal model can only acquire extra supervision signals from informal sentences. Because the main objective of FST is the informal$\to$formal transfer, the additional informal sentences were not well utilized in previous studies. In addition, these semi-supervised models incorporate many auxiliary modules such as style discriminators, to achieve state-of-the-art results, which result in rather complicated frameworks and more model parameters.

As noisy informal sentences are easier to acquire from online resources, we attempt to take a different view from existing approaches, by adopting additional \textit{source}-side (informal) sentences via SSL. We gain insights from the state-of-the-art approaches for semi-supervised image and text classification \citep{sohn2020fixmatch,xie2020unsupervised,berthelot2019mixmatch,zhang2021flexmatch,chen-etal-2020-mixtext} and propose a simple yet effective SSL framework for FST using purely informal sentences. Our approach employs \textit{consistency training} to generate pseudo-parallel data from additional informal sentences. Specifically, we enforce the model to generate similar target sentences for
an unlabeled source-side sentence and its perturbed version, making the model more robust against the noise in the unlabeled data. In addition, a supervised loss is trained simultaneously to transfer knowledge from the clean parallel data to the unsupervised consistency training.

Data perturbation is the key component of consistency training and significantly affects its performance. To obtain a successful SSL framework for FST, we first empirically study the effects of various data perturbation approaches. Specifically, we explore easy data augmentation methods, such as random \textit{word deletion}, and advanced data augmentation methods, such as \textit{back-translation}. We also handcraft a line of rule-based data perturbation methods to simulate the features of informal sentences, such as \textit{spelling error injection}. Furthermore, we propose three data filtering approaches in connection with the three evaluation metrics of FST: style strength, content preservation, and fluency. Specifically, we adopt \textit{style accuracy}, \textit{source}-BLEU, and \textit{perplexity} as three metrics to filter out low-quality pseudo-parallel data based on a threshold. We also propose a dynamic threshold algorithm to automatically select and update the thresholds of source-BLEU and perplexity.

We evaluate our framework on the two domains of the GYAFC benchmark: \textit{Entertainment \& Music} (E\&M) and \textit{Family \& Relationships} (F\&R). We further collect 200k unpaired informal sentences for each domain to perform semi-supervised training. Experimental results verify that our SSL framework can enhance the performance of the strong supervised baseline, a pretrained T5-large \citep{2020t5} model, by a substantial margin, and improve the state-of-the-art results by over 2.0 BLEU scores on both GYAFC domains.  Empirically, we also deduce that simple word-level data augmentation approaches are better than advanced data augmentation methods that excessively alter the sentences, and \textit{spelling error injection} is especially effective. In addition, our evaluation-based data filtering approach can further improve the performance of the SSL framework. Furthermore, we also conduct low-resource experiments by reducing the size of parallel data. Surprisingly, our framework could achieve the state-of-the-art results with only less than 40\% of parallel data, demonstrating the advantage of our method in low-resource situations.

\section{Related Work}

\begin{figure*}[ht]
\setlength{\belowcaptionskip}{-.4cm}
\small
    \centering
    \includegraphics[]{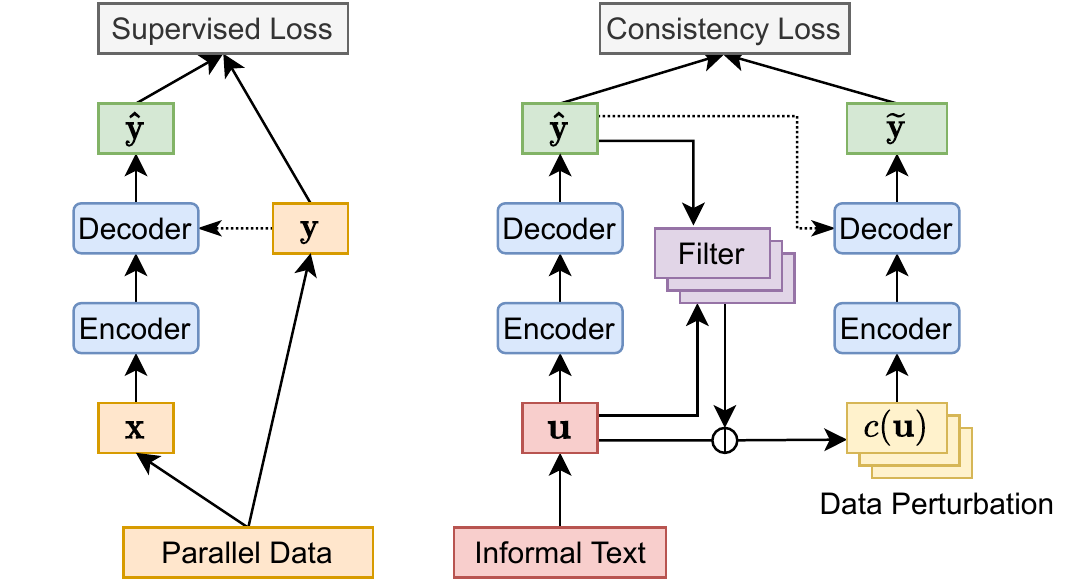}
    \caption{Overview of our semi-supervised consistency training framework which jointly optimizes two losses: (1) a supervised loss trained on parallel data; (2) a consistency training loss, where the model first generates a pseudo target $\hat{\mathbf{y}}$ for an additional informal text, then it is combined with a perturbed input $c(\mathbf{u})$ to train the model if passing the data filter.
    The dotted arrows indicate teacher forcing in the encoder-decoder model.}
    \label{framework}
\end{figure*}
\paragraph{Formality style transfer} FST is an important branch of text style transfer. For FST, \citet{rao2018dear} released a high-quality parallel dataset - GYAFC, comprising two sub-domains and approximately 50k parallel data for each domain. Previous studies \citep{rao2018dear,niu2018multi, xu2019formality,zhang2020parallel} typically train seq2seq encoder-decoder models on this benchmark. Recent studies \cite{harnessing, cari, chawla-yang-2020-semi, lai-etal-2021-thank} have deduced that fine-tuning large-scale pretrained models such as GPT-2 \citep{gpt2} and BART \citep{bart} on the parallel corpora can improve the performance.
To address the data-scarcity problem of parallel datasets,  \citet{zhang2020parallel} proposed three data augmentation techniques to augment pseudo-parallel data for training. Similar to prior research on text style transfer that adopt back-translation  \citep{zhang2018style, lample2018multiple, prabhumoye2018style, Luo19DualRL}, some other approaches on FST \citep{shang2019semi, xu2019formality, chawla-yang-2020-semi} adopt a cycle-reconstruction scheme, where an additional backward transfer model is jointly trained together with the forward transfer model, and the two models generate pseudo-paired data for each other via iterative back-translation. Although \citet{xu2019formality} and \citet{chawla-yang-2020-semi} train a single model to perform bidirectional transfer, the generation of both directions remain disentangled by a control variable, making each direction rely on the unlabeled data of its target side. Therefore, the unlabeled informal sentences exert no direct effects on the informal$\to$formal transfer. In contrast, our work focuses on how to better utilize source-side unlabeled data (i.e., informal sentences) using SSL and does not introduce any extra models.

\paragraph{SSL with consistency regularization} SSL is popular for its advantage in utilizing unlabeled data. Consistency regularization (also known as consistency training) \citep{sajjadi2016regularization} is an important component of recent SSL algorithms on image and text classification \citep{miyato2018virtual,tarvainen2017mean,berthelot2019mixmatch, sohn2020fixmatch}. It enforces a model to produce invariant predictions for an unlabeled data and its perturbed version. These studies developed different data perturbation \citep{xie2020unsupervised, berthelot2019mixmatch} or data filtering \citep{zhang2021flexmatch,xu2021dash} approaches to improve the performance. However, few studies have been made on how to apply consistency training in natural language generation (NLG) tasks such as FST because of the different target spaces, i.e., instead of single class labels or probabilities, the output of NLG is the combination of discrete NL tokens. This renders the experiences in classification tasks not applicable to FST. For instance, classification probabilities are typically adopted as the metric to filter high-confidence pseudo-examples for consistency training in classification tasks~\citep{sohn2020fixmatch,xie2020unsupervised,zhang2021flexmatch}, which is implausible in FST. A similar study~\cite{he2019revisiting} improved self-training by injecting noise into unlabeled inputs and proved its effectiveness on machine translation and text summarization; however, self-training involves multiple iterations to collect pseudo-parallel data and retrain the model, hence the training is not end-to-end. 
In this study, we explore various data perturbation strategies and propose effective data filtering approaches to realize a successful consistency-based framework for FST, which may also provide useful insights for future studies on semi-supervised NLG.

\section{Method}

\subsection{Base Model}
FST involves rewriting an informal sentence into a formal one.  Formally, given a sentence $\mathbf{x}=(x_1, x_2, \dots, x_n)$ of length $n$ with style $S$, our objective is to transform it into a target sentence $\mathbf{y}=(y_1,y_2,\dots,y_m)$ of length $m$ and style $T$, while preserving its content.

Following prior studies \citep{rao2018dear,zhang2020parallel, chawla-yang-2020-semi,lai-etal-2021-thank} on FST, we employ the supervised baseline as a seq2seq encoder-decoder model that directly learns the conditional probability $P(\mathbf{y}|\mathbf{x})$ from parallel corpus $\mathcal{D}$ comprising $(\mathbf{x}, \mathbf{y})$ pairs. The objective is the cross-entropy loss between the decoder outputs and the ground-truth target sentences:

\begin{align}
\nonumber
    \mathcal{L}_{sup} &= \mathbb{E}_{(\mathbf{x},\mathbf{y})\sim D}[-\log P(\mathbf{y}|\mathbf{x}; \theta)]  \\ 
    &= \mathbb{E}_{(\mathbf{x},\mathbf{y})\sim D}[-\sum_i \log P(y_i|y_{1:i-1}, \mathbf{x}; \theta)],
    \label{eq:sup}
\end{align}
where $\theta$ denotes the model parameters.

\subsection{Consistency Training}
Our approach leverages the idea of consistency regularization \citep{sajjadi2016regularization} and enforces a model to generate similar target sentences for an original and perturbed unlabeled sentence. Simultaneously, the model is also trained on the supervised data. Accordingly, the knowledge garnered from supervised training can be gradually transferred to unsupervised training. An overview of our framework is presented in Figure \ref{framework}. Typically, the consistency training loss is computed on the divergence between predictions on an unlabeled input $\mathbf{u}$ and its perturbed version $\Tilde{\mathbf{u}}=c(\mathbf{u})$, where $c(\cdot)$ is the perturbation function and $\mathbf{u}\in \mathcal{U}_S$ represents a source-side unlabeled sentence (in our case, an informal sentence). Formally, consistency training can be defined as minimizing the following unsupervised loss:
\begin{align}
    \mathbb{E}_{\mathbf{u}\sim \mathcal{U}_S} D\left [P(\mathbf{y}|\mathbf{u}; \theta)||P(\mathbf{y}|c(\mathbf{u}); \theta)\right ],
\end{align}
where $D[\cdot || \cdot]$ denotes a divergence loss. In practice, we adopt pseudo-labeling~\cite{lee2013pseudo} to train the unsupervised loss, for which we fix the model parameter $\theta$ to predict a ``hard label'' (pseudo target sentence) $\hat{\mathbf{y}}$ for $\mathbf{u}$ and enforce the consistency of model prediction by training $\theta$ with $(c(\mathbf{u}), \hat{\mathbf{y}})$. Hence the unsupervised objective can be optimized as a standard cross-entropy loss as follows:
\begin{align}
    \mathcal{L}_{unsup}=\mathbb{E}_{\mathbf{u}\sim \mathcal{U}_S}\mathbb{E}_{\hat{\mathbf{y}}\sim P(\mathbf{y}|\mathbf{u};\hat{\theta})}[-\log P(\hat{\mathbf{y}}|c(\mathbf{u});\theta)],
    \label{eq:unsup}
\end{align}
where $\hat{\theta}$ denotes a fixed copy of $\theta$.
This training process does not introduce additional model parameters. The entire additional training cost to supervised learning is a training pass and a generation pass for each unlabeled sentence.

As the overall objective, we train a weighted sum of the supervised loss in Equation \eqref{eq:sup} and the unsupervised loss in Equation \eqref{eq:unsup}:
\begin{align}
    \mathcal{L}=\mathcal{L}_{sup} + \lambda\mathcal{L}_{unsup},
\end{align}
where $\lambda$ represents a hyper-parameter for balancing the effects of supervised and unsupervised training. To achieve a good initial model for consistency training, we first pretrain the model on the supervised loss for several warm-up steps.

\subsection{Data Perturbation Strategies}
\label{sec:da}
Data perturbation is the key component of consistency-based SSL algorithms \citep{xie2020unsupervised,chen-etal-2020-mixtext} and significantly affects the performance. In this section, we briefly introduce a collection of different data perturbation methods explored in this research.

First, we consider some easy data augmentation methods commonly used for supervised data augmentation, which includes
\begin{itemize}
    \item \textbf{word deletion (drop)\footnote{We abbreviate each method for ease of denotation.}}: to randomly drop a proportion of words in the sentence.
    \item  \textbf{word swapping (swap)}: to randomly swap a proportion of words with their neighbouring words.  
    \item \textbf{word masking (mask)}: to randomly replace words with a mask token ``\_''.
    \item \textbf{word replacing with synonym (synonym)}: to randomly replace some words with a synonym based on WordNet \citep{wordnet}.
\end{itemize}

In addition, we consider advanced data augmentation methods that have proven effective in semi-supervised text classification \citep{xie2020unsupervised}:
\begin{itemize}
    \item \textbf{back-translation}: to translate a sentence into a pivot language, then translate it back to obtain a paraphrase of the original one.
    \item \textbf{TF-IDF based word replacing (tf-idf)}: to replace uninformative words with low TF-IDF scores while retaining those with high TF-IDF values.
\end{itemize}

Furthermore, we handcraft a set of rule-based data perturbation for FST. There are some typical informal expressions in the parallel corpus, such as the use of slang words and abbreviations, capitalized words for emphasis, and spelling errors. Some existing studies \cite{harnessing, cari} adopt editing rules to revise such informal expressions as a preprocessing step. Inspired by these, we propose the adoption of opposite rules to synthesize such noises. We consider the following methods:
\begin{itemize}
    \item \textbf{spelling error injection (spell)}: to randomly inject spelling errors to a proportion of words by referring to a spelling error dictionary.
    \item  \textbf{word replacing with abbreviations (abbr)}: to replace all the words in the sentence with their abbreviations or slang words (e.g., ``are you'' $\to$ ``r u'') by referring to an abbreviation dictionary.
    \item  \textbf{word capitalization (capital)}: to randomly capitalize a proportion of words.
\end{itemize}
  These rule-based methods can inject noise into the unlabeled informal sentences without changing its informality, but strengthening it instead.

\subsection{Evaluation-Based Data Filtering}
In the consistency training loss, the noisy pseudo-target $\hat{\mathbf{y}}$ is generated from the decoder model and may exert negative effects on the training. Therefore, we propose three evaluation-based data filters in connection with the evaluation metrics of FST.

Specifically, we attempt to measure the quality of pseudo-target sentences by considering the three most important evaluation criteria of text style transfer: \textbf{style strength}, \textbf{content preservation}, and \textbf{fluency}. Next, we comprehensively explain each evaluation metric and the corresponding data filter.

\textbf{Style strength} measures the formality of generated sentences. Typically, people adopt binary classifiers such as TextCNN \citep{textcnn} classifiers to judge the formality of a sentence \citep{lai-etal-2021-thank}. Inspired by this,
we pretrain a TextCNN formality classifier on the parallel training corpus (i.e., GYAFC) to distinguish between informal and formal sentences. For an unlabeled informal sentence $\mathbf{u}$ and its pseudo target sentence $\hat{\mathbf{y}}$, we maintain $(c(\mathbf{u}),\hat{\mathbf{y}})$ for unsupervised training only when 
\begin{align}
    p_{cls}^+(\hat{\mathbf{y}}) - p_{cls}^+(\mathbf{u}) > \sigma,
\end{align}
where $p_{cls}^+(\cdot)$ represents the probability of the sentence being formal, predicted by the style classifier and $\sigma$ is a threshold of the probability. This guarantees that only the sentence pairs with strong style-differences are used for consistency training.

\textbf{Content preservation} is another important evaluation metric of FST, typically measured with BLEU between the ground-truth target sentence and the model generations. In unsupervised text style transfer where no ground-truth target exists, \textit{source}-BLEU is adopted as an alternative, i.e., the BLEU scores between the source input sentence and the generated target sentence.
Similarly, we propose the adoption of \textit{source}-BLEU between $\mathbf{u}$ and $\hat{\mathbf{y}}$ as the metric to filter out pseudo targets that present poor content preservation. 

\textbf{Fluency} is also used to evaluate the quality of generated sentences. We follow \citep{hu2020text} to pretrain an N-gram language model on the training data to estimate the empirical distributions of formal sentences. Then, the \textit{perplexity} score is calculated for the pseudo target sentence $\hat{\mathbf{y}}$ by the language model. The motivation is that the sentences with lower perplexity scores match the empirical distribution of formal sentences better, and are thus considered as more fluent.

A natural idea is to filter out pseudo-parallel data based on a \textit{source}-BLEU or a \textit{perplexity} threshold. However, it is infeasible to determine the optimal threshold for the two metrics beforehand because the pseudo paired data are generated on-the-fly during the training and we cannot know the distribution of the BLEU or perplexity scores. In addition, choosing the BLEU/perplexity threshold is not as easy as tuning the style probability $\sigma$ because they heavily depend on the data distribution and exhibit varying ranges of values.

\subsection{Dynamic Threshold Selection}

To realize the selection of thresholds for the BLEU- and perplexity- based filters, we propose a dynamic threshold strategy based on the distribution of the scores computed for already generated pseudo-paired sentences. Specifically, we maintain an ordered list $L$ to store the scores calculated for previously generated pseudo data and update it continuously following the training. At each iteration, a batch of new scores are inserted into $L$ while maintaining the decreasing order of the list. Subsequently, we update the threshold as the value at a certain position $L[\phi \times len(L)]$ in the score list, where $len(L)$ denotes the length of the current score list and $\phi\in [0, 1]$ represents a ratio that determines the threshold's position in the list. We only keep pseudo data with scores higher (or lower for perplexity scores) than the threshold for consistency training. This actually makes $\phi$ approximately the proportion of pseudo data we keep for training, making it more convenient to control the trade-off between the qualities and quantities of selected pseudo data. More details are provided in Appendix \ref{sec:formal}, \ref{sec:thresh}.

\section{Experiments}
We introduce the experimental settings in Section \ref{sec:setup}. To obtain relevant findings on how to build an effective consistency training framework for FST, we first empirically study the effects of multiple data perturbation methods in Section \ref{sec:dp} and prove the effectiveness of consistency training via comparisons with the base model. Then, we validate our consistency training model with different data filtering methods in Section \ref{sec:df} and demonstrate their additional effects on the SSL framework. Based on the findings in these two experiments, we further compare our best models with previous state-of-the-art models in Section \ref{sec:main}. We also include case studies in Section \ref{sec:main} to present some qualitative examples. Finally, we conduct low-resource experiments (Section \ref{sec:low}) to demonstrate our method's advantage when less parallel data are available.

\subsection{Experimental Settings}
\label{sec:setup}

\begin{table}[t]
    \centering
    \small
    \begin{tabular}{|c|c|c|c|c|}
    \hline
     Dataset    & Train & Val & Test & Unlabeled \\
     \hline
    E\&M      &  52595& 2877& 1416 & 200k\\
    \hline
    F\&R  &51967 &2788& 1432 & 200k\\
    \hline
    \end{tabular}
    \caption{The statistics of datasets.}
    \label{tab:dataset}
\end{table}

\begin{table*}[t]
    \centering
 \small
    \begin{tabular}{lcccccc}
        \toprule
        & \multicolumn{3}{c}{\textbf{E\&M}} & \multicolumn{3}{c}{\textbf{F\&R}} \\
 \cmidrule(lr){2-4} \cmidrule(lr){5-7} 
     Model variants    & BLEU & Acc(\%)& HM & BLEU & Acc(\%) & HM \\
     \midrule
      base model  & 76.87  & 90.04 & 82.94 & 80.32 & 84.01 & 82.12 \\
      no-perturbation & 76.41 &	88.49 & 	82.01 & 79.22 &	84.46 &	81.75 \\	
      \midrule
      drop & 77.55 & 93.15 & 84.64& 80.53 &	86.56 &	83.44 \\
      swap & 77.90 &	93.43&	84.96& 81.07 &	85.96 & 83.44 \\
      mask & 77.52&	93.93&	84.94 & 80.69 &	86.41 &	83.45\\
      synonym &77.48 & 93.64 & 84.80 & 80.49 &	86.26 &	83.28 \\
      \midrule
      back-translation & 76.07 &	90.11&	82.50& 79.96 &	84.91 &	82.36 \\
      tf-idf & 76.89 & 92.58 &84.01 & 80.48 &	\textbf{86.94} &	83.58 \\
      \midrule
      abbr & 77.55 &	93.64 &	84.84 & 81.00 &	\textbf{86.94} &	\textbf{83.86} \\
      capital &77.54 &	93.15 &	84.63 & 80.74 &	85.74 &	83.16\\
      spell & \textbf{78.37} &\textbf{94.21}&	\textbf{85.56}& \textbf{81.09} &	85.59 &	83.28  \\
        \bottomrule
    \end{tabular}
    \caption{Effects of different data perturbations in our approach on the test splits of GYAFC. The best scores among all the model variants are boldfaced.}
    \label{tab:da-res}
\end{table*}

\paragraph{Datasets} We evaluate our framework on the GYAFC~\cite{rao2018dear} benchmark for formality style transfer. It comprises crowdsourced informal-formal sentence pairs split into two domains, namely, E\&M and F\&R. The informal sentences in the dataset were originally selected from the same domains in Yahoo Answers L6 corpus\footnote{https://webscope.sandbox.yahoo.com/catalog.php
?datatype=l}. We focus on the \textit{informal-formal} style transfer because it is more realistic in applications.
We further collected massive amounts of informal sentences from each of the two domains in Yahoo Answers L6 corpus as the unsupervised data. The statistics of the datasets are presented in Table \ref{tab:dataset}.

\begin{table*}[t]
    \centering
    \small
    \begin{tabular}{lcccccc}
        \toprule
        & \multicolumn{3}{c}{\textbf{E\&M}} & \multicolumn{3}{c}{\textbf{F\&R}} \\
 \cmidrule(lr){2-4} \cmidrule(lr){5-7} 
     Model variants    & BLEU & Acc(\%)& HM & BLEU & Acc(\%) & HM \\
    
      \midrule
      spell (no-filter)& 78.37 & 94.21 &85.56 & 81.09 &	85.59 &	83.28  \\
      spell (+style) & 78.19 &	93.79 &	85.28 &\textbf{81.37} &	\textbf{86.41} &	\textbf{83.81 }\\
      spell (+bleu) & \textbf{78.75} &	\textbf{94.56} &	\textbf{85.94} &\textbf{81.11 }&	\textbf{86.34} &	\textbf{83.64 }\\
      spell (+lm) & 78.24 &	\textbf{94.56 }&	\textbf{85.63} & 80.93 &	\textbf{86.34} &	\textbf{83.55}\\
        \bottomrule
    \end{tabular}
    \caption{Effects of different data filtering methods in our approach on the test splits of GYAFC. Scores larger than the no-filter variant are in \textbf{bold}.}
    \label{tab:filter-res}
\end{table*}
\paragraph{Implementation Details} We employ PyTorch \citep{pytorch} for all the experiments.  We pretrain a TextCNN style classifier on the supervised data for each domain of GYAFC, following the setting in \cite{lai-etal-2021-thank}.
The same classifier is adopted for both the style accuracy evaluation and the style strength filter in our SSL framework. We adopt HuggingFace Transformers \citep{transformers} library's implementation of pretrained T5-Large \citep{2020t5} as the base model. We adopt the Adam \citep{adam} optimizer with the initial learning rate $2\times 10^{-5}$ to train all the models.  More details of hyper-parameters and model configurations are provided in Appendix \ref{sec:des}.

\paragraph{Evaluation Metrics}
The main evaluation metric for FST is the BLEU score between the generated sentence and four human references in the test set. We adopt the corpus BLEU in NLTK~\cite{loper-bird-2002-nltk} following \citep{chawla-yang-2020-semi}. In addition, we also pretrained a TextCNN formality classifier to predict the formality of transferred sentences and calculate the accuracy (Acc.). Furthermore, we compute the harmonic mean of BLEU and style accuracy as an overall score, following the settings in \cite{lai-etal-2021-thank}.

\subsection{Effects of Data Perturbation Methods}
\label{sec:dp}
In this experiment, we validate the effectiveness of our consistency training framework and compare the effects of different data perturbation methods. Specifically, we adopt the nine data perturbation methods introduced in Section \ref{sec:da} and include the \textit{no-perturbation} variant that indicates directly using an unlabeled sentence and its pseudo target to train the unsupervised loss. We adopted no data filtering strategy in this experiment to simplify the comparison. 

As shown in Table \ref{tab:da-res}, our framework could consistently improve the base model by using different perturbation methods; however, back-translation resulted in mostly lower results than the base model. This contradicts the conclusion in \cite{xie2020unsupervised} that back-translation is especially powerful for semi-supervised text classification. We attribute this to the fact that back-translation tends to change the entire sentence into a semantically similar but syntactically different sentence. Compared with other word-level perturbation strategies, back-translation triggers a larger mismatch between the perturbed input and the pseudo-target sentence generated from the unperturbed input, leading to a poor content preservation ability of the model. In contrast, simple word-level noises achieved consistently better results, especially spell error (\textit{spell}), random word swapping (\textit{swap}), and abbreviation replacing (\textit{abbr}). These three methods tend to alter the words but do not lose their information while other methods eliminate the entire word by deleting (\textit{drop, mask}) or replacing it with another word (\textit{synonym, tf-idf}). This may also cause a larger mismatch between the pseudo input and output. 

Hence, we draw the conclusion that \textit{simple word-level perturbations tend to bring more effects}. This differs from the observations in text classification~\cite{xie2020unsupervised} because content preservation is important in FST. In particular, we also found that \textit{spell} achieved the highest BLEU scores on both datasets. However, adding no perturbation even resulted in a worse performance than the base model. Moreover, \textit{capital} is also relatively weaker than the other two rule-based methods because it only changes the case of a chosen word. This suggests that the perturbation should not be too simple either.

\begin{table*}[t]
    \centering
    \small
    \resizebox{\textwidth}{!}{
    \begin{tabular}{lccccccc}
        \toprule
        &  & \multicolumn{3}{c}{\textbf{E\&M}} & \multicolumn{3}{c}{\textbf{F\&R}} \\
\cmidrule(lr){3-5}   \cmidrule(lr){6-8}  
     Models   &  unlabeled data & BLEU & Acc(\%)& HM & BLEU & Acc(\%) & HM \\
     \midrule
      NMT\citep{rao2018dear} & no & 68.41 & - & - & 74.22 & - & - \\
  
      Hybrid Annotations$^\dagger$$^*$ \citep{xu2019formality} & yes & 69.28  & 89.83 & 78.23 & 74.36 & 82.96 & 78.42 \\
     NMT-Multi-task$^\dagger$ \citep{niu2018multi} & no & 72.01  & 88.84 & 79.54 & 75.35 & 80.03 & 77.62 \\
     GPT-CAT \citep{harnessing} & no & 71.39  & - & - & 77.26 & - & - \\
     Transformers (DA) \citep{zhang2020parallel}& yes &  74.24&- &- &77.97 & -& -\\
     CARI \citep{cari}& no & 74.31  & - & - & 78.05 & - & - \\
      Chawla's$^\dagger$ \citep{chawla-yang-2020-semi} & yes & 76.17  & 91.88 & 83.29 & 79.92 & 83.63 & 81.73 \\
     BART-large+SC+BLEU$^\dagger$ $^*$ \citep{lai-etal-2021-thank}& no& 76.50 & 94.42 & 84.52 & 79.25  & \textbf{90.69}& \textbf{84.58}\\
       \midrule
      Ours (base) & no & 76.87  & 90.04 & 82.94 & 80.32 & 84.01 & 82.12 \\
       Ours (best) & yes & \textbf{78.75}  & \textbf{94.56} & \textbf{85.94}& \textbf{81.37} & 86.41 & 83.81 \\

        \bottomrule
    \end{tabular}}
    \caption{Comparison between our approach and existing works on the test splits of GYAFC. $^\dagger$ indicates we recalculate the scores with our evaluation metrics for the output given in the paper. Otherwise, we copy the results from the paper. $^*$ indicates that the model used training data from both domains and is not comparable to our model.  }
    \label{tab:res}
\end{table*}

\subsection{Effects of Data Filtering}
\label{sec:df}
In this section, we analyze whether our proposed data filters are beneficial to the performance of our consistency training framework. Specifically, we chose the most effective data perturbation method \textit{spell} to analyze the effects of adding the three data filters: style strength (\textit{style}), content preservation (\textit{bleu}), and fluency (\textit{lm}) filters. As presented in Table \ref{tab:filter-res}, the results for different datasets and different filters have different tendencies. For example, adding the \textit{style} filter on the E\&M dataset caused negative effects while contributing the best results to the F\&R domain.

Although a filter does not necessarily improve the result, this is reasonable because filters result in less pseudo data for model training and it is difficult to control the trade-off between the quality and the quantity of selected data. Nevertheless, we still observe that the \textit{bleu} filter contributes to the highest performance of \textit{spell} for all the metrics on the E\&M domain, while \textit{style} benefits the performance of \textit{spell} the most on F\&R, leading to the best performing models of our approach\footnote{Empirically, we also found that mixing up three filters achieved no better results than a single filter, possibly because this filtered out too much pseudo data.}.

\begin{table}[t]
    \centering
    \resizebox{\columnwidth}{!}{
    \begin{tabular}{lcccccc}
    \toprule
            
    & \multicolumn{2}{c}{Formality} & \multicolumn{2}{c}{Fluency} &\multicolumn{2}{c}{Meaning} \\
    \cmidrule(lr){2-3}\cmidrule(lr){4-5}\cmidrule(lr){6-7}
     Model & E\&M & F\&R & E\&M & F\&R & E\&M & F\&R \\
     \midrule
    Chawla's & 1.46  &1.22 &1.85  &1.80 &\textbf{1.87} &1.88\\
    Ours (base) &1.42 & 1.28& 1.84 &1.82 & 1.86 & \textbf{1.95} \\ 
    Ours (best) &\textbf{1.55} &\textbf{1.41} & \textbf{1.88}& \textbf{1.85} &\textbf{1.87} & 1.88 \\ \bottomrule
    \end{tabular}}
    \caption{Human evaluation results.}
    \label{tab:human}
\end{table}

\begin{table}[t]
    \centering
    \resizebox{\columnwidth}{!}{
    \begin{tabular}{lcccc}
        \toprule
        & \multicolumn{2}{c}{\textbf{E\&M}} & \multicolumn{2}{c}{\textbf{F\&R}} \\
 \cmidrule(lr){2-3} \cmidrule(lr){4-5} 
     \#Parallel data   & BLEU & Acc(\%) & BLEU & Acc(\%)  \\
     \midrule
     100 (base) & 59.94 &	61.58 &	 65.13 &	49.32 	 \\ 
      
     100 (ours) & \textbf{64.40}	&\textbf{82.91} & \textbf{71.11} &\textbf{55.11} \\
     \midrule
     1000 (base) & 70.49 &	83.26 &	 75.36 &	\textbf{76.58}\\
     1000 (ours) & \textbf{72.22 }&	\textbf{85.81}  & \textbf{76.70} &	76.20 \\
     \midrule
     5000 (base) & 75.13 &	\textbf{89.55} &	 77.65 &	78.38 \\ 
     5000 (ours) & \textbf{75.67 }&	87.08 	& \textbf{78.87} &\textbf{81.01}	\\
     \midrule
     20000 (base) & 76.55 &	90.96  & 79.25 &	83.33 \\
     20000 (ours)& \textbf{76.59} &	\textbf{92.09} &	 \textbf{80.61} &\textbf{86.11} 	 \\
        \bottomrule
    \end{tabular}}
    \caption{Experimental results on test sets under low-resource settings with varied parallel data size.}
    \label{tab:few-shot}
\end{table}

\label{sec:case}

\begin{table*}[t]
  \small
    \centering{
    \begin{tabular}{lll}
        \toprule
      \multirow{5}{*}{Example 1} & Source  & \textit{I like natural / real girls, I don't like fake looking prissy drama queens.}\\ 
    &   Ours(best)  &  \textit{I like natural looking girls, not pretentious drama queens.}\\
    &   Ours(base) &  \textit{I like natural, real girls, I do not like fake looking, prissy drama queens.}\\
    &   Chawla’s &  \textit{I like natural and real girls , I do not like fake looking prissy drama queens .}\\
    &   Human-Annotation &  \textit{I like natural and real girls, not fake-looking, prissy drama queens.}\\

       \midrule
  \multirow{5}{*}{Example 2}  &    Source  & \textit{That's like Broke Back Mountain for little John Wanye.}\\ 
   &    Ours(best)  &  \textit{That is similar to ``Broke Back Mountain'' for John Wayne.}\\
    &   Ours(base) &  \textit{That is like ``Broke Back Mountain'' for John Wayne.}\\
   &    Chawla’s &  \textit{That is like Broke Back Mountain for little John Wanye .}\\
     &  Human-Annotation &  \textit{That is similar to ``Brokeback Mountain'' for young John Wayne.}\\
       \midrule
 \multirow{5}{*}{Example 3}   &    Source  & \textit{You guys don't have any reason to hate each other.}\\ 
    &   Ours(best)  &  \textit{You do not have any reason to hate each other.}\\
    &   Ours(base) &  \textit{You guys do not have any reason to hate each other.}\\
    &   Chawla’s &  \textit{You guys do not have any reason to hate each other .}\\
     &  Human-Annotation &  \textit{There is no reason for you two to dislike each other.}\\
       \bottomrule
    \end{tabular}}
    \caption{Examples sampled from the test set outputs.}
    \label{tab:case_study}
\end{table*}

\subsection{Comparison with Previous Works}
\label{sec:main}
We compare our best model with the following previous studies on GYAFC. 
\begin{itemize}
    \item \textbf{NMT} \citep{rao2018dear} is an LSTM-based encoder-decoder model with attention. 
    \item \textbf{GPT-CAT} \citep{harnessing} adopts GPT-2 and rule-based pre-processing for informal sentences. 
    \item \textbf{NMT-Multi-task} \citep{niu2018multi} jointly solves monolingual formality transfer and formality-sensitive machine translation via multi-task learning.
    \item \textbf{Hybrid Annotations} \citep{xu2019formality} trains a CNN discriminator in addition to the transfer model and adopts a cycle-reconstruction loss to utilize unsupervised data.
    \item \textbf{Transformers (DA)} \citep{zhang2020parallel} uses three data augmentation methods, including back-translation, formality discrimination, and multi-task transfer. 
    \item \textbf{CARI} \citep{cari} improves GPT-CAT by using BERT \citep{devlin2018bert} to select optimal rules to pre-process the informal sentences. 
    \item \textbf{Chawla's} \citep{chawla-yang-2020-semi} uses language model discriminators and maximizing mutual information to improve a pretrained BART-Large \citep{bart} model, along with a cycle-reconstruction loss to utilize unlabeled data.
    \item \textbf{BART-large+SC+BLEU} \citep{lai-etal-2021-thank} improves BART-large by incorporating reinforcement learning rewards to enhance style change and content preservation.
\end{itemize}
We also report the results of \textbf{Ours (base)}, our backbone T5-large model, and \textbf{Ours (best)}, our best performing models selected from Table \ref{tab:filter-res}. 

As observed in Table \ref{tab:res}, \textbf{Ours (best)} outperforms previous state-of-the-art models by a substantial margin and improves the BLEU scores from 76.17 and 79.92 to 78.75 and 81.37, respectively, on the E\&M and F\&R domains of the GYAFC benchmark. Although \textbf{BART-large+SC+BLEU} achieved better results on the Acc. of F\&R, the only released official outputs of \textbf{BART-large+SC+BLEU} were obtained from a model that was trained on the training data of both domains and adopted rewards to directly optimize style accuracy; hence, it is not directly comparable to our model. \textbf{Ours (best)} improves the fine-tuned T5-large baseline by a large margin as well, demonstrating the effectiveness of our SSL framework.

\paragraph{Human Evaluation} We also conduct human evaluation to better capture the quality of the models' outputs. Following \citep{zhang2020parallel}, we measure the \textit{Formality}, \textit{Fluency}, and \textit{Meaning Preservation} of generated sentences by asking two human annotators to assign a score ranging from \{0, +1, +2\} regarding each aspect. We randomly sampled 50 examples from the test set of each domain and compare the generated outputs of \textbf{Ours (base)}, \textbf{Ours (best)}, and the previous state-of-the-art \textbf{Chawla's} model trained on the single-domain data. In addition, the annotators were unaware of the corresponding model of each output. As shown in Table \ref{tab:human}, the human evaluation results are consistent with the automatic evaluation results: \textbf{Ours (base)} is competitive compared with \textbf{Chawla's}, while \textbf{Ours (best)} improves over the base model and outperforms the previous state-of-the-art on all the metrics, except that it presents lower results on \textit{Meaning} than \textbf{Ours (base)} on F\&R. More details on human evaluation can be found in Appendix \ref{sec:human}.

\paragraph{Qualitative Examples} We present some of the generated outputs of \textbf{Ours (base)}, \textbf{Ours (best)}, and \textbf{Chawla's} in Table \ref{tab:case_study}. It can be observed that all the models can produce high-quality outputs with considerable formality, meaning preservation and fluency. Nevertheless, \textbf{Ours (best)} exhibits a stronger capability to modify the original sentence, especially for some informal expressions, leading to the best performance on the \textit{Formality} metric. For example, it replaced ``like'' with ``similar to'' in Example 2 and deleted the informal word ``guys'' in Example 3. However, it may alter the original sentence so much that the meaning of the sentence is changed to some extent (Example 1). This may explain why \textbf{Ours (best)} achieves a lower \textit{Meaning} score than \textbf{Ours (base)} on F\&R.

\subsection{Low-Resource Experiments}
\label{sec:low}
We also simulate the low-resource settings by further reducing the size of available parallel data. Specifically, we randomly sample from the original training data with a size in the range of \{100, 1000, 5000, 20000\} and compare the results of the base model T5-Large with our SSL model. The size of unlabeled data remains 200k for each domain. We adopt the \textit{spell} data perturbation without any data filter and avoid exhaustive hyper-parameter tuning. Table \ref{tab:few-shot} demonstrates that our framework is especially effective under few-shot settings when only 100 parallel data are available. By comparing with previous state-of-the-art results on FST, we can observe that our approach can achieve competitive results with only 5000 ($<10\%$) parallel training data, and even better results with only 20000 ($<40\%$) parallel examples.

\section{Conclusion}
In this study, we proposed a simple yet effective consistency-based semi-supervised learning framework for formality style transfer. Unlike previous studies that adopted cycle-reconstruction to utilize additional target-side sentences for back-translation, our method offers a different view, to leverage source-side unlabeled sentences. Without introducing additional model parameters, our method can easily outperform the strong supervised baseline and achieve the new state-of-the-art results on formality style transfer datasets. For future work, we will attempt to generalize our approach to other text generation scenarios.

\section*{Acknowledgements}
This paper is based on results obtained from a project, JPNP18002, commissioned by the New Energy and Industrial Technology Development Organization (NEDO). Ao Liu acknowledges financial support from the Advanced Human Resource Development Fellowship for Doctoral Students, Tokyo Institute of Technology.

\bibliography{anthology,custom}
\bibliographystyle{acl_natbib}


\appendix

\section{Detailed Experimental Settings}
\label{sec:des}

\subsection{Hyper-Parameters}
We set the max length of input sentences to 50 Byte-Pair Encoding \citep{sennrich2016neural} tokens. The weight of unsupervised loss $\lambda$ is set to 1.0 in all our experiments, which is an empirical choice from previous studies~\cite{sohn2020fixmatch}. The batch size is 8 for the supervised objective and 56 for the unsupervised objective, such that the model can leverage more unlabeled data for training.  The threshold $\sigma$ for the style strength filter is set to 0.8 and the threshold ratio $\phi$ is set to 0.4 for both the content preservation and fluency filters. We tested $\sigma$ in the discrete range between 0.5 and 0.9 and for $\phi$, we searched over the values between 0.1 and 0.8. Although the chosen values of $\sigma$ and $\phi$ are not necessarily the best for all the datasets, we fix them in later experiments for their reasonable results.

\subsection{Training Details}
We train two binary style classifiers on each domain of GYAFC. The training data are the formal and informal sentences in the original training sets of the E\&M and F\&R domain. The classifiers are validated on the formal sentences in the original validation set. The classifier for E\&M 
could achieve 95.69\% accuracy on the validation set, while the classifier for F\&R achieved 94.70\%.
We adopt a 4-gram Kneser-Ney language model to compute perplexity scores for the fluency data filter.
During semi-supervised training, we first pretrain the model solely on the supervised data for 2000 steps to achieve a good initialization of the model parameters. Then, we jointly train the supervised and consistency losses simultaneously. The model checkpoint is validated with an interval of 1000 steps and selected based on the best BLEU score on the validation set. Early stopping is also adopted with patience 10.
We employ beam search with beam width 5 for the model's generations and pseudo-target prediction\footnote{The pseudo target can also be obtained by sampling methods.}.
All our experiments are conducted on NVIDIA A100 (40GB) GPUs. 

\subsection{Details of Unlabeled Data Collection}
We collected 200k from each of the E\&M and F\&R domains of Yahoo Answers L6 corpus. The collection procedure is as follows. (1) We chose the passages labeled ``<bestanswer>'' in the corpus and tokenized them into separate sentences. (2) We filtered out sentences with formality scores larger than 0.5 (i.e. judged as formal) predicted by the style classifier we built for model evaluation. (3) We built an N-gram language model by training on the informal sentences in the original training data of GYAFC, and used it to generate perplexity scores for these sentences. We kept 200k sentences with lowest perplexity scores, such that we obtained a collection of the most informal sentences in the corpus. We only observed one overlapping sentence with the test set of each domain, which we considered negligible and kept in the data.

\subsection{Details of Data Perturbation}
\label{sec:ddp}
All our data perturbation methods are implemented based on the nlpaug\footnote{https://github.com/makcedward/nlpaug} library. For the \textit{abbr} perturbation, we adopt the abbreviation dictionary which \citet{xformal} used for rule-based pre-processing.
 We set the ratio of perturbed words in a sentence to 0.1 for all word-level perturbation methods and deduced that increasing the ratio could often result in lower results, as that will enhance the difference between the original and perturbed sentences, which is consistent with our conclusion in Section \ref{sec:dp}. We present examples of all data perturbation methods in Table \ref{tab:exp-dp}.

We also attempted mixing different perturbations with \textit{spell}, but did not obtain better results than single \textit{spell}. This can also be attributed to the conclusion that simple perturbations are even better.

\begin{table*}[t]
\small
    \centering
    \begin{tabular}{|c|c|}
    \hline
    Original sentence & \textit{Well first you have to get lots of hands on experience.}\\ \hline
    Word deletion & \textit{Well first you have to get lots of on experience.} \\ \hline
    Word swapping     & \textit{Well first \textcolor{red}{have you} to get lots of hands on experience. } \\\hline
    Word masking  &\textit{ Well first \textcolor{red}{\_} have to get lots of hands on experience.}\\ \hline
    Word replacing with synonym & \textit{Well first you have to \textcolor{red}{begin} lots of hands on experience.} \\ \hline
    Back-translation & \textit{\textcolor{red}{well} first you have to get lots of \textcolor{red}{years} on experience.}\\  \hline
    TF-IDF based word replacing & \textit{Well first you have \textcolor{red}{walmartmusic} get lots of hands on experience }\\ \hline
    Spelling error injection & \textit{Well first you have to get lots of hands \textcolor{red}{or} experience.}\\ \hline
    Word replacing with abbreviations & \textit{Well first \textcolor{red}{u} have to get lots of hands on experience.} \\ \hline
    Word capitalization & \textit{Well \textcolor{red}{FIRST} you have to get lots of hands on experience.}\\
    \hline
    
    \end{tabular}
    \caption{Examples of data perturbation methods. Different words compared to the original sentence are marked as red.}
    \label{tab:exp-dp}
\end{table*}

\section{Formal Description of the Algorithm}
\label{sec:formal}
Here, we provide a formal algorithmic description of our consistency training framework in Algorithm \ref{alg} and assume that we adopt content preservation (BLEU) data filtering or fluency (perplexity) data filtering in this algorithm to include the formal description of our dynamic threshold strategy. We omit the case when we adopt style strength filtering because it does not use the dynamic threshold and is more straightforward to understand.
\begin{algorithm*}[ht]
\small
  \caption{Training Procedure of our approach using dynamic threshold selection}
  \label{alg}
\begin{algorithmic}[1]
\State{\bfseries Input:} Parallel corpus $\mathcal{D}=\{\mathbf{x},\mathbf{y}\}^M$, unlabeled corpus of source-side sentences $\mathcal{U}_S=\{\mathbf{u}\}^N$, initialized model parameters $\theta$; perturbation function $c(\cdot)$, supervised batch size $B$, unsupervised batch size $\mu B$, weight factor $\lambda$, filter type $ft \in$ \{BLEU, perplexity\}, a data filter score function $f$, an decreasing-ordered score list $L$, a function $len(\cdot)$ that returns the length of a list.

{\Comment{Warm-up training}}
\State Initialize $\theta$ with pretrained T5.
\State Finetune $\theta$ on $\mathcal{D}$ via Equation \eqref{eq:sup}.

{\Comment{Semi-supervised training}}
\Repeat
\State Sample a batch $\mathcal{B}_D = \{(\mathbf{x}_i, \mathbf{y}_i)\}_{i=1}^B$ from $\mathcal{D}$.

\State Sample a batch $\mathcal{B}_U = \{\mathbf{u}_i\}_{i=1}^{\mu B}$ from $\mathcal{U}_S$.

\State Obtain $ \mathcal{B}_U^{'}=\{\Tilde{\mathbf{u}}_i |\Tilde{\mathbf{u}}_i=c(\mathbf{u}_i)\}_{i=1}^{\mu B}$.
\State Generate pseudo targets $\mathcal{B}_Y = \{\hat{\mathbf{y}}_i|\hat{\mathbf{y}}_i=\text{argmax} P(\mathbf{y}| \mathbf{u}_i; \theta)\}_{i=1}^{\mu B}$.
\State Compute a batch of data filter scores $L_B=\{b_i| b_i=f(\mathbf{u}_i, \hat{\mathbf{y}}_i)\}_{i=1}^{\mu B}$
\State Insert $L_B$ into $L$ while maintaining the decreasing order of $L$.
\State Obtain $s = L[\sigma \times len(L)]$ as the threshold.
\If{$ft=$ BLEU}
  \State Obtain a filtered pseudo-parallel batch $\mathcal{B}_f=\{(\Tilde{\mathbf{u}}_i, \hat{\mathbf{y}}_i)| b_i > s, i = 1, ,\dots,\mu B\}$
\ElsIf{$ft=$ perplexity}
\State Obtain a filtered pseudo-parallel batch $\mathcal{B}_f=\{(\Tilde{\mathbf{u}}_i, \hat{\mathbf{y}}_i)| b_i < s, i = 1, ,\dots,\mu B\}$
\EndIf

\State Compute consistency loss $ \mathcal{L}_{unsup}=\mathbb{E}_{(\Tilde{\mathbf{u}},\hat{\mathbf{y}})\sim \mathcal{B}_f}[-\log P(\hat{\mathbf{y}}|\Tilde{\mathbf{u}};\theta)]$.
\State Compute supervised loss $ \mathcal{L}_{sup} = \mathbb{E}_{(\mathbf{x},\mathbf{y})\sim \mathcal{B}_D}[-\log P(\mathbf{y}|\mathbf{x}; \theta)]$.
\State Optimize $\mathcal{L}=\mathcal{L}_{sup} + \lambda\mathcal{L}_{unsup}$ and update $\theta$.

\Until CONVERGE
\end{algorithmic}
\end{algorithm*}

\section{Details of Dynamic Threshold Selection}
\label{sec:thresh}
Here, we provide more details of the dynamic threshold strategy for the content preservation and fluency filters. In practice, we do not filter any pseudo data in the initial warm-up steps of consistency training, to initialize the score list. Furthermore, after iterating an epoch of the unsupervised data, we keep the current threshold fixed and do not update the score list any more. The score list is implemented as a skiplist to enable $O(\log N)$ insertion into an ordered list. The overall time complexity of the data filtering is $O(\log 1 + \log 2 + \cdots + \log N)=O(\log N!) = O(N\log N)$, where $N$ is the number of unlabeled data.

\section{Details of Human Evaluation }
\label{sec:human}
We describe the rating criteria in the human evaluation. We ask two well-educated annotators to rate the \textit{formality}, \textit{fluency}, and \textit{meaning preservation} on a discrete scale from 0 to 2 for the model outputs, following \cite{zhang2020parallel}. During the annotation, we randomly shuffle the sentences from the three models and make the model names invisible to annotators.

\paragraph{Formality} The annotator are asked to rate the formality change level given a source informal sentence and the generated output sentence, regardless of the fluency and meaning preservation. If the output sentence improves the formality of the source sentence significantly, the score will be 2 points. If the output sentence improves the formality but still keeps some informal expressions, or the improvement is minimal, it will be rated 1 point. If there is no improvement on the formality, it will be rated 0 points.

\paragraph{Fluency} The fluency is rated 2 points if the output sentence is meaningful and has no grammatical error. If the target sentence is meaningful but contains some minor grammatical errors, it will be rated 1 point. If the sentence is incoherent, it will be rated 0 points.

\paragraph{Meaning Preservation} Given a source sentence and a corresponding output sentence, the raters are asked to ascertain how much information is preserved in the output sentence compared to the input sentence. If the two sentences are exactly equivalent, the output obtains 2 points. If they are mostly equivalent but different in some trivial details, the output will receive 1 point. If the output omits important details that alter the meaning of the input sentence, it is rated 0 points.

\end{document}